\documentclass[10pt,twocolumn,letterpaper]{article}

\usepackage[pagenumbers]{cvpr} 

\usepackage{times}
\usepackage{epsfig}
\usepackage{graphicx}
\usepackage{amsmath}
\usepackage{amssymb}
\usepackage{multirow}
\usepackage{multicol}
\usepackage{algorithm}
\usepackage{algorithmic}
\usepackage{booktabs}
\usepackage{mathrsfs}
\usepackage{colortbl}  
\usepackage{array}   
\usepackage{overpic}
\usepackage{microtype}      

\makeatletter
\newcommand\tabcaption{\def\@captype{table}\caption}
\newcommand\figcaption{\def\@captype{figure}\caption}
\makeatother

\usepackage[dvipsnames]{xcolor}

\definecolor{cvprblue}{rgb}{0.21,0.49,0.74}
\definecolor{Gray}{gray}{0.86}
\usepackage{hyperref}
\hypersetup{breaklinks,colorlinks,citecolor=cvprblue}

\title{Regressor-Segmenter Mutual Prompt Learning for Crowd Counting}

\author{Mingyue Guo\textsuperscript{1,2}, 
Li Yuan\textsuperscript{2,3}, 
Zhaoyi Yan\textsuperscript{2}\thanks{Corresponding author.},
Binghui Chen,
Yaowei Wang\textsuperscript{2},
Qixiang Ye\textsuperscript{1,2}\\
\textsuperscript{1}University of Chinese Academy of Sciences\quad \textsuperscript{2}Pengcheng Lab\quad \textsuperscript{3}Peking University\\
{\tt\small guomingyue21@mails.ucas.ac.cn}\quad{\tt\small yuanli-ece@pku.edu.cn}\quad{\tt\small chenbinghui@bupt.cn}\\ \quad{\tt\small yanzhaoyi@outlook.com}\quad{\tt\small wangyw@pcl.ac.cn}\quad{\tt\small qxye@ucas.ac.cn}
}

\begin{document}
\maketitle

\begin{abstract}

Crowd counting has achieved significant progress by training regressors to predict instance positions. In heavily crowded scenarios, however, regressors are challenged by uncontrollable annotation variance, which causes density map bias and context information inaccuracy.
In this study, we propose mutual prompt learning (mPrompt), which leverages a regressor and a segmenter as guidance for each other, solving bias and inaccuracy caused by annotation variance while distinguishing foreground from background.
In specific, mPrompt leverages point annotations to tune the segmenter and predict pseudo head masks in a way of point prompt learning.
It then uses the predicted segmentation masks, which serve as spatial constraint, to rectify biased point annotations as context prompt learning. 
mPrompt defines a way of mutual information maximization from prompt learning, mitigating the impact of annotation variance while improving model accuracy.
Experiments show that mPrompt significantly reduces the Mean Average Error (MAE), demonstrating the potential to be general framework for down-stream vision tasks.
Code is enclosed in the appendix.

\end{abstract}

\section{Introduction}
\label{sec:Intro}
Crowd counting, which estimates the number of people in images of crowded or cluttered backgrounds, has garnered increasing attention for its wide-ranging applications in public security~\cite{kang2018beyond,onoro2016towards}, traffic monitoring~\cite{guerrero2015extremely}, and agriculture~\cite{aich2017leaf,lu2017tasselnet}. 
Many existing methods converted crowd counting as a density map regression problem~\cite{lempitsky2010learning,zhang2015cross,sam2017switching,li2018csrnet}, $i.e.$, generating density map targets by convolving the point annotations with the predefined Gaussian kernels and then training a model to learn from these targets. 

\begin{figure}[t]
    \centering
    \includegraphics[width=1\linewidth]{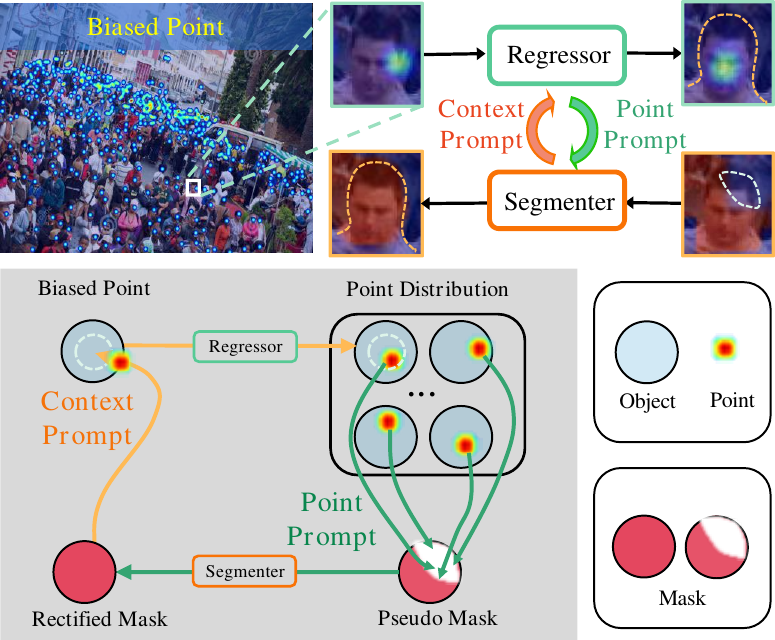}
    \caption{\textbf{Upper: The biased point annotation impedes accurate model learning.} mPrompt leverages context prompt and point prompt to mine spatial context and rectify biased annotation for crowd counting. \textbf{Lower: Illustration of mutual prompt learning (mPrompt)}, which completes pseudo segmentation mask by using point prompt learning. Meanwhile, it leverages the rectified masks as spatial context information to refine biased point annotations in a way of context prompt learning. (Best viewed in color) }
    \label{fig:mot_method}
\end{figure}

Unfortunately, point annotations exhibit considerable variances, termed label variance, which impedes the accurate learning of models. As shown in Fig.~\ref{fig:mot_method}, label variance is an inherent issue, where the annotated point are coarsely placed within head regions rather than at precise center positions. 
To mitigate the label variance, loss relaxation approaches~\cite{ma2019bayesian,Wan2020modeling, wang2020DMCount} modified the strict pixel-wise loss constraint via constructing probability density functions.
Segmentation-based approaches~\cite{modolo2021understanding, zhao2019leveraging, shi2019counting} suppressed background noises by introducing an auxiliary branch to regressor networks~\cite{modolo2021understanding}.

Unfortunately, loss relaxation methods comprise point position variance, which could introduce background noises to the regressor. Segmentation-based methods manage to alleviate label variance using spatial context, but are challenged by the inaccurate context information. To obtain accurate context information while alleviating background noises in a systematic framework remains to be elaborated.

In this study, the pivotal question we seek to address is: \emph{How to obtain precise spatial context information to alleviate the impact of label variance for crowd counting?} We propose a simple-yet-effective mutual prompt learning framework, Fig.~\ref{fig:mot_method}, which leverages a regressor and a segmenter as guidance for each other. This framework comprises a head segmenter, a density map regressor, and a mutual learning module. Specifically, mPrompt leverages the point annotation to tune a segmenter and predict pseudo head masks in a way of point prompt learning. As illustration in Fig.~\ref{fig:mot_method}, the point prompt provides statistical distribution (random and uncertain locations) of points to refine the object mask. The objective of the segmenter is to isolate head regions, so as to learn comprehensive and accurate pseudo segmentation masks. Such pseudo segmentation masks are treated as spatial context to rectify biased point annotations in a way of context prompt learning. 
This mutual prompt process fosters information gain between the segmenter and the density map regressor, driving them to enhance each other and ultimately reach an optimal state. 

The contributions of this study are summarized as follows:
\begin{itemize}
\item {
We propose a mutual prompt learning (mPrompt) framework, which incorporates a segmenter and a regressor and maximizes their complementary for crowd counting. To our best knowledge, this is the first attempt to unify learning accurate context information and alleviating background noises using mutual prompt.
}

\item {We design feasible point prompt by unifying the predicted density map with the ground-truth one, and plausible mask prompt by unifying/intersecting the predicted density map with a segmentation mask.}

\item Experiments
conducted on the popular crowd-counting datasets, including ShanghaiTechA/B~\cite{zhang2016single}, UCF-QNRF~\cite{wang2020nwpu} and NWPU~\cite{idrees2018composition} demonstrate mPrompt's effectiveness when addressing label variance. Particularly, mPrompt achieves new state-of-the-art performances on multiple benchmark datasets.
\end{itemize}
\section{Related Work}
\label{sec:related}
\textbf{Density Regression Method.} Nowadays, density map regression~\cite{lempitsky2010learning} is widely used in crowd counting~\cite{zhang2016single,sam2017switching,li2018csrnet,cao2018scale,liu2019adcrowdnet,liu2019crowd,zhang2019relational,bai2020adaptive,song2021sasnet,wan2021generalized,song2021rethinking,wang2021uniformity} due to its simple and effective learning strategies.
Nevertheless, many density regression approaches neglected scale variation of heads, and thereby is challenged by the inconsistency between density maps and features caused by labeling variance.

To tackle scale variance, multi-scale feature fusion layers~\cite{jiang2019crowd,song2021sasnet,jiang2019learning}, attention mechanisms~\cite{liu2019adcrowdnet, zhang2019relational,jiang2020attention,lin2022boosting,dong2022clrnet}, perspective information~\cite{shi2019revisiting,yan2019persp,yang2020reverse,yan2021crowd}, and dilated networks~\cite{bai2020adaptive,yan2019persp} were proposed. 

To mitigate the side effect of inaccurate point annotations, distribution matching~\cite{wang2020DMCount,lin2021direct}, generalized localization loss~\cite{wan2021generalized}, and density normalized precision~\cite{song2021rethinking} are proposed to minimize the discrepancy between the predicted maps and point annotations.
For so many approaches proposed, however, density regression remains challenged by the label variance issue, which is expected to be tackled by introducing segmentation-based context information.

\textbf{Segmentation-based Method.}
In early years, Chan et al.~\cite{chan2008privacy} and Ryan et al.~\cite{ryan2009crowd} proposed to segment foreground objects to distinct clusters, and regress the features of each cluster to determine the overall object counts.
Recent studies~\cite{zhao2019leveraging, shi2019counting, modolo2021understanding} began to incorporate image segmentation as an auxiliary task to leverage spatial context information while mitigating the effects of false regression.
These methods typically utilized the coarse ``ground-truth" segmentation maps, which are simply derived from the noisy point annotation maps. As a result, they lack robust and precise spatial information, and are prone to label variability. In contrast, this study smoothly acquires precise spatial information about head positions, reducing label variance through the deployment of mutual prompt learning.
The significant advantage of our approach upon conventional segmentation-based approaches lies in that it can fully explore the statistical distribution (random and uncertain locations) of points to refine the object mask in a way of point prompt learning.

\textbf{Prompt Learning.}
In the era of large language models~\cite{Devlin_Chang_Lee_Toutanova_2019,brown2020language}, prompt learning has been shown to be a powerful tool for solving various natural language processing (NLP) tasks~\cite{Liu_Yuan_Fu_Jiang_Hayashi_Neubig_2022,radford2019language,brown2020language}. various prompt learning strategies including prompt engineering~\cite{lu2021fantastically,brown2020language}, prompt assembling~\cite{jiang2020can}, and prompt tuning~\cite{ouyang2022training}, are respectively proposed. 
Inspired by the success of prompt learning in NLP, vision prompt learning approaches~\cite{jia2022visual,bar2022visual} are proposed. 
The challenge lies in how to design plausible prompts which can guide and enhance the learning of models on specific tasks.

In this study, we take a further step to mutual prompt learning, with the aim to enhance both the regression and segmentation models in a unified framework.
While the term ``prompt'' typically refers to ``guidance/hint'' embedding into the pretrained large model in the forward process, our work extends its application to the realm of backward gradient propagation (via point and context prompt in this paper).
We also extend our method by integrating pre-trained large-scale models, capitalizing on their extensive knowledge base. This integration enables our model to achieve robust performance while maintaining parameter efficiency during training.

%
\begin{figure*}[t]
\centering

\begin{overpic}[scale=0.9]{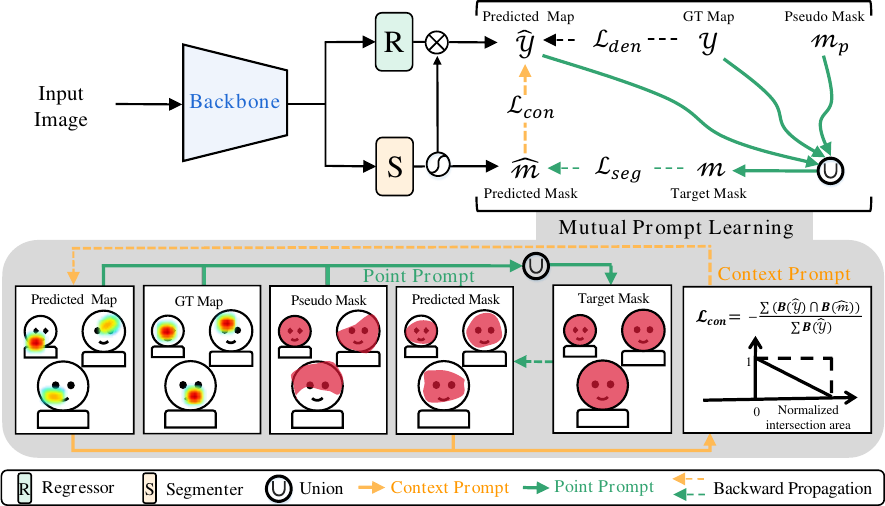}

\end{overpic}
\scriptsize
\caption{mPrompt consists of four components: a shared backbone for feature extraction, a regressor for density map ($\hat{y}$) prediction, a segmenter for head region ($\hat{m}$) estimation, and a mutual prompt learning module.
}\label{fig:arch}
\end{figure*}

\section{The Proposed Approach}
\label{sec:method}
The proposed approach integrates a regressor and a segmenter for density map and segmentation mask prediction. In what follows, we first unify the segmenter with a regressor to construct a two-branch network. We then introduce mutual prompt learning to the network, which encompasses point prompts given by the regressor and context prompts provided by the segmenter.

\subsection{Unifying Segmenter with Regressor}\label{sec:usg}

\textbf{Network Architecture.} As shown in Fig.~\ref{fig:arch}(upper), mPrompt consists of a shared CNN backbone, a density regressor $\mathcal{R}$ and a head segmenter $\mathcal{S}$, which are trained in an end-to-end fashion. The shared backbone is derived from a HRNet by truncating layers from \textit{stage4}~\cite{wang2020deep}. 
To seamlessly unify the segmenter with the regressor, a self-attention module applied to them to enhance features of the regressor, Fig.~\ref{fig:arch}. Denoting $\mathcal{S}(x)$ and $\mathcal{R}(x)$ as the features of the regressor and the segmenter for an input image ${x}$, the self-attention operation is applied on $\mathcal{S}(x)$ and $\mathcal{R}(x)$ as $Sigmoid\big(\mathcal{S}(x)\big) \bigotimes \mathcal{R}(x)$, where $\bigotimes$ is the element-wise multiplication. 
With feature self-attention, the regressor preliminarily incorporates the context information provided by the segmenter.

%
\begin{figure*}[t]
\centering
\begin{overpic}[scale=0.22, tics=2]{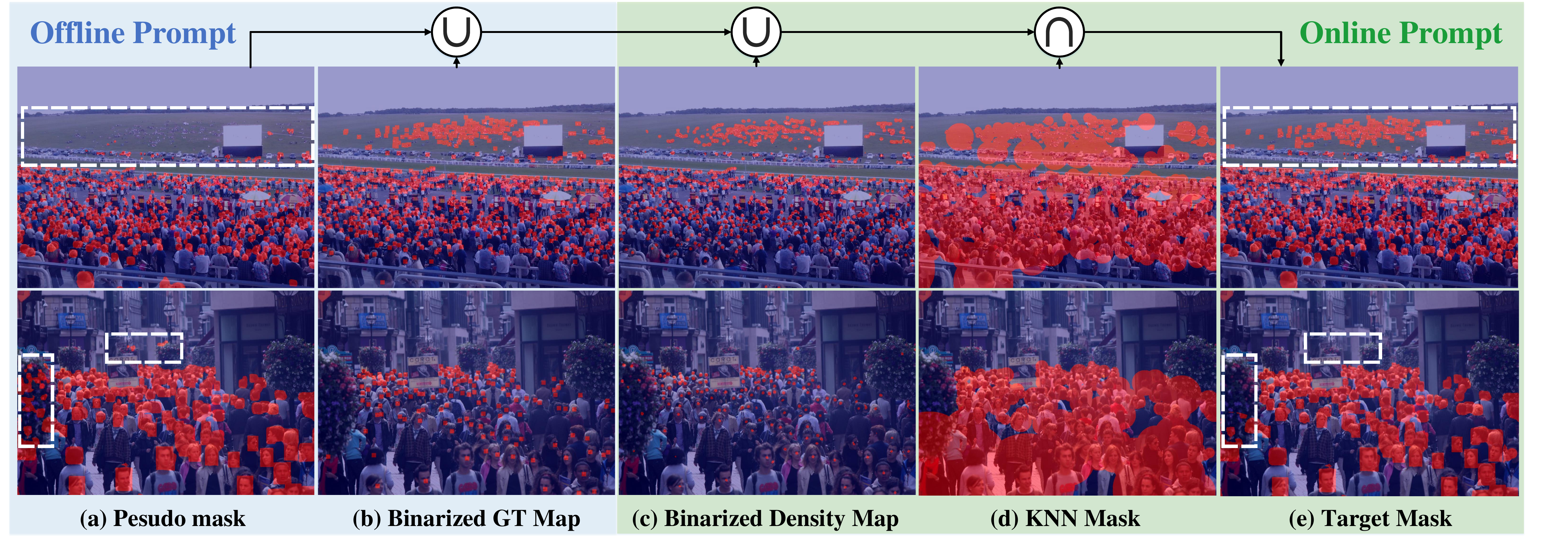}
\end{overpic}
\vspace{-1em}
\caption{Illustration of the generation of prompt information for the segmenter. White boxes highlight key regions for better clarity. The red-shaded areas represent the head segmentation mask, demonstrating the pseudo mask's inaccuracy when compared to the more precise updated target mask. With offline prompt, the prompted segmenter tends to predicted more complete head regions but unfortunately introduces background noises. With online prompt, background noises are reduced. (Best viewed in color with zoom)}
\label{fig:offline_refinement}
\end{figure*}

\begin{figure}[t]
    \centering
    \includegraphics[width=1\linewidth]{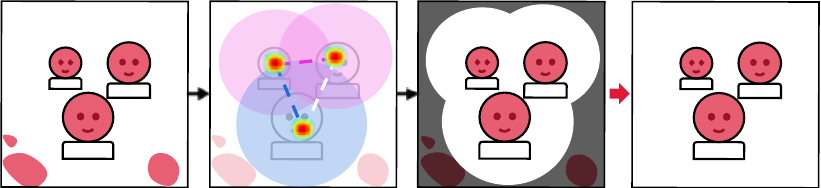}
    \caption{Illustration of $K$-NN algorithm, which removes background noises from the target segmentation mask. }
    \label{fig:knn}
\end{figure}

The regressor predicts the density map $\boldsymbol{\hat{y}}$ for the input image $\boldsymbol{x}$, and the segmenter predicts the head mask $\boldsymbol{\hat{m}}$. The regressor and segmenter are designed using an identical architecture, comprising Conv-BN-ReLU blocks. Specifically, three Conv-BN-ReLU blocks are adopted to decrease the feature channel size progressively from $128$ to $64$, and eventually down to $32$ followed by a self-attention operation. A convolution layer of kernel size $1$ followed by ReLU/Sigmoid layer squeezes the features to density/segmentation maps.

 {\textbf{Segmenter Learning.}}
Each point annotation is expanded to a density map ($\boldsymbol{y}$) and a target mask ($\boldsymbol{m}$), which however are noisy and inaccurate. Fortunately, existing datasets, such as NWPU~\cite{wang2020nwpu}, provide point and box annotations, which can be expanded to pseudo masks for segmenter training. A point pseudo mask is derived by applying dilation to the point density map, which are converted to a segmentation mask after binarization. Following~\cite{zhao2019leveraging, shi2019counting, modolo2021understanding}, we train the segmenter using the cross-entropy loss function $\mathcal{L}_{s}$ defined on point pseudo masks. 

As elucidated by experiments, using point-based pseudo masks to train a segmenter exhibits a challenge in assimilating spatial information. This limitation primarily stems from the fact that the learning targets for both the segmenter and regressor are manually created from dot annotations, which intrinsically do not convey any spatial information. 
To develop an advanced segmenter, we further leverage the box annotations provided by the NWPU dataset~\cite{wang2020nwpu}. A box pseudo mask is produced by attributing values of $1$ to locations within the heads and $0$ to the background. Accordingly, the overall loss for the regressor and segmenter is defined as $\mathcal{L} =  \mathcal{L}_{den}+\lambda_{s}\mathcal{L}_{seg}$ where $\lambda_{s}$ is a parameter to balance the two losses~\footnote{Please refer to the appendix for details of training a segmenter using point and box/pseudo masks.}. 

\subsection{Segmenter Learning with Point Prompt}
\label{sec:seg_learning_with_pp}
As shown in Fig.~\ref{fig:arch}, point prompt defines a procedure to  refine the target mask $\boldsymbol{m}$ using the pseudo mask ${m}_{p}$, the ground-truth density map $\boldsymbol{y}$ and the predicted density map $\boldsymbol{\hat{y}}$. In specific, we utilize the pseudo mask $\boldsymbol{m}_{p}$ (offline obtained via a segmenter pretrained on NWPU box annotations) and ground-truth density map ($\boldsymbol{y}$) for offline prompt, and the density map $\boldsymbol{\hat{y}}$ for online prompt, which guarantees the renewal of the segmentation map via the prompt from the regressor. 
When training the segmenter, the binary cross entropy loss is applied.

{\textbf{Offline Prompt.}}
This is performed by unifying the segmentation pseudo mask $\boldsymbol{m}_{p}$ with the binarized ground-truth density map $\boldsymbol{y}$, as
\begin{equation}
    \boldsymbol{m} = \boldsymbol{m}_{p} \cup \mathbf{B}({\boldsymbol{y}}),
\label{eq:point_prompt_off}
\end{equation}
where $\cup$ denotes the union operation performing pixel-wise OR operation between two matrices. $\mathbf{B}(\cdot)$ defines a binarization function: the density map is binarized with a $0$ threshold to form a mask. 
Supervised by training targets $\boldsymbol{m}$ from all the training images, the segmenter tends to absorb the distribution (random and uncertain locations) of points. After prompt learning, the prompted segmenter tends to predict more complete head regions (the top row of Fig.~\ref{fig:offline_refinement}) where the initialized segmenter fails to predict. 

{\textbf{Online Prompt.}}
With the offline prompt, the accuracy of the predicted density map can be improved after $\kappa$ epochs of training, so that it can be used to improve the target segmentation mask.
Following the initial $\kappa$ training epochs, $\hat{y}$ should possess credibility and aid in introducing reliable distributions (Gaussian blobs randomly situated around the point annotations) of head regions. 
As a result, integrating $\hat{y}$ into point prompt learning further assists in predicting the comprehensive head regions.
Meanwhile, the union operation defined in offline/online prompt inevitably introduce background noises from the density map to the target mask. 
To solve, we further leverage a $K$-NN algorithm to filter out background noises (the bottom row of Fig.~\ref{fig:offline_refinement}) at the end of online prompt, which defined as interaction operation. Online prompt defines the following union and intersection operations, as
\begin{equation}
   \boldsymbol{m} \gets (\boldsymbol{m} \boldsymbol \cup (\mathbf{B}({\boldsymbol{\hat{y}}})) \boldsymbol \cap \boldsymbol{m}_{K},
\label{eq:point_prompt_on}
\end{equation}
where $m_{K}$ is a context mask defined by a spatial $K$-NN algorithm applied on the point annotations~Fig.~\ref{fig:knn}. In specific, for a point annotation, the spatial $K$-NN algorithm finds its $K$ nearest point annotations. The minimum circle area covering the $K$ nearest point annotations is defined as the context mask $m_{K}$.

\subsection{Regressor Learning via Context Prompt}
\label{sec:regressor_segmenter}
With point prompt, the segmenter absorbs distribution of the annotated points so that it produces more accurate mask predictions. Such mask predictions serve as a spatial information to improve the regressor in turn, which is termed as context prompt. 
In specific, the context prompt is defined as a constraint, which encourages the predicted density map $\boldsymbol{\hat y}$  falling the target mask $\boldsymbol{m}$. This is implemented by introducing a context prompt loss to the framework, as
\begin{equation}
    \boldsymbol{\mathcal{L}}_{con}(\boldsymbol{\hat y}, \boldsymbol{\hat m}) = -
    \frac{ \Sigma{\big(\mathbf{B}({\boldsymbol{\hat y}})} \cap \mathbf{B}({\boldsymbol{\hat m}}) \big) } {\Sigma{\mathbf{B}({\boldsymbol{\hat y}})}},   
\label{eq:con_prompt}
\end{equation}
where $\Sigma$ accumulates the values of all pixels. To minimize the context prompt 
loss, intersection term in Eq.~\ref{eq:con_prompt} must be large, which implies the prediction $\boldsymbol{\hat y}$ of the regressor falling in the predicted mask area ($\boldsymbol{\hat m}$) of the segmenter. In other words, the segmenter serves as the context prompt of the regressor.
When training the regressor, the conventional MSE constrain is defined as the density map construction loss.

\subsection{Mutual Prompt Learning}
Given the point prompt defined by Eq.~\ref{eq:point_prompt_off} and Eq.~\ref{eq:point_prompt_on}, and the context prompt defined by Eq.~\ref{eq:con_prompt}, the mutual prompt learning is performed in an end-to-end fashion by optimizing the following loss function, 
\begin{equation}
    \boldsymbol{\mathcal{L}} = \lambda_{d}\mathcal{L}_{den}(\boldsymbol{\hat y}, \boldsymbol{y}) + \lambda_{s}\mathcal{L}_{seg}(\boldsymbol{\hat m}, \boldsymbol{m}) + \lambda_{c}\mathcal{L}_{con}(\boldsymbol{\hat y}, \boldsymbol{\hat m}),
\label{}
\end{equation}
where $\lambda_{d}$, $\lambda_{s}$ and $\lambda_{c}$ are experimentally defined regularization factors.

In summary, our mPrompt comprises three components: (1) With point prompt learning, the segmenter absorbs statistical distribution (random and uncertain locations) of points to predict more accurate target masks. (2) With context prompt learning, the predicted density map is constrained to fall into the target mask regions, which in turn improve the density regression. (3) Unifying point prompt learning with context prompt learning in a framework with shared backbone and training the network parameters in an end-to-end fashion create mutual prompt learning.

\subsection{Extension to Foundation Model}
\label{sec:extension} 
%

\begin{figure}[t]
    \centering
    \includegraphics[width=1\linewidth]{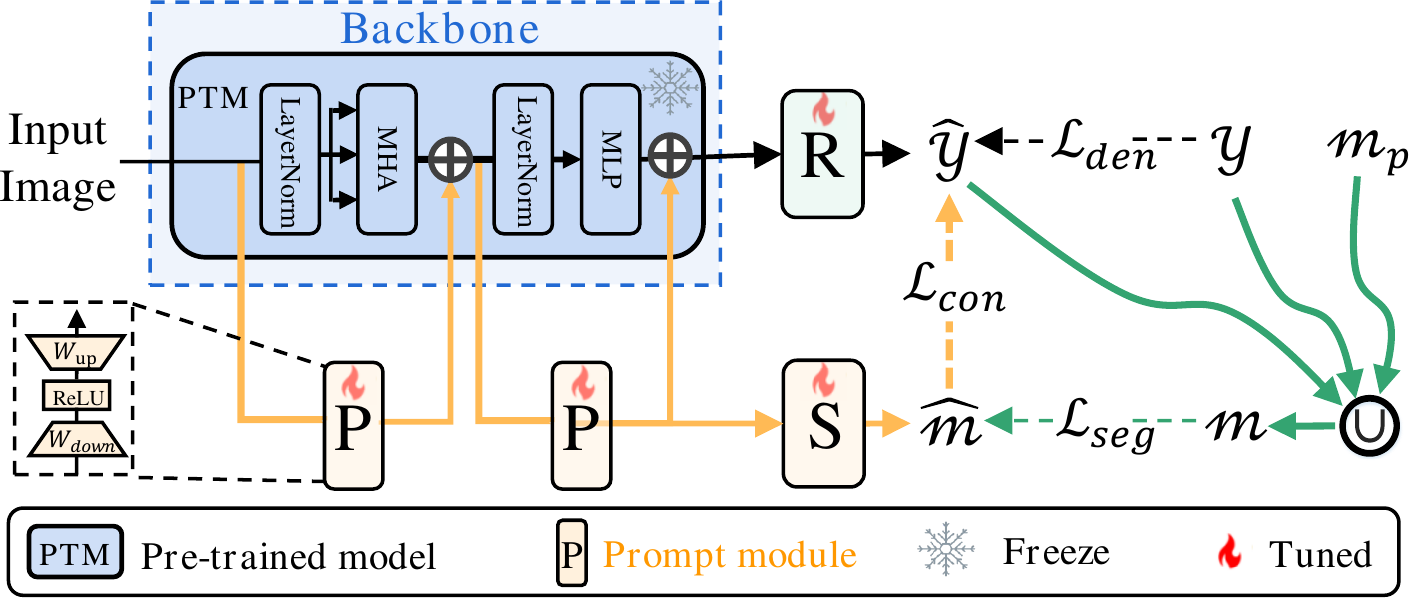}
    \caption{mPrompt with learnable prompt modules based on pre-trained model.}
    \label{fig:extension}
\end{figure}

Our mPrompt approach can be further applied to foundation model adaptation. This involves expanding the context prompt into a feature insertion strategy, which enhances the utilization of the extensive knowledge embedded in pre-trained large models, as demonstrated in Fig.~\ref{fig:extension}. 
In this process, the context prompt is modulated by learnable prompt modules. Such prompt modules are implemented using adapter mechanism\footnote{Please refer to the appendix for more details.}~\cite{DBLP:conf/icml/HoulsbyGJMLGAG19}. 
Our primary goal is to integrate comprehensive context information into foundational models, specifically for crowd counting. This aims to make effective use of the representational knowledge in pre-trained large models by only fine-tuning a small number of parameters.

During the inference phase, the learnable prompt modules, along with the backbone and regressor, are retained, while the segmenter branch is discarded. These prompt modules function as context prompts, facilitating the insertion of features into the backbone.
\subsection{Interpretive Analysis}
The proposed approach is justified from the perspective of mutual information~\cite{ahn2019variational}.
mPrompt can be generally interpreted as a procedure to maximize the mutual information $\mathcal{I}$ of a regressor ($f^r$) and a segmenter ($f^s$). 
The point prompt is interpreted as
\begin{equation}
   \mathbf{H}(f^s,f^r) = \mathbf{H}(f^s|f^r) + \mathbf{H}(f^r)
\label{eq:point_prompt}
\end{equation}
where $\mathbf{H}(\cdot)$ is information entropy, the $\mathbf{H}(f^s|f^r)$ is conditional and the $\mathbf{H}(f^s,f^r)$ is joint entropy. 
Denote the parameters of the model as $\theta$. To minimize $\boldsymbol{\mathcal{L}}_{con}(\boldsymbol{\hat y}, \boldsymbol{\hat m})$ is equivalent to maximize $\log\frac{ \Sigma{\big( \mathbf{B}({\boldsymbol{\hat y}})} \cap \mathbf{B}({\boldsymbol{\hat m}}) \big) } {\Sigma{\mathbf{B}({\boldsymbol{\hat y}})}}$. Then the context prompt is interpreted as
%
\begin{equation}
\begin{split}
\arg\max_{\theta} &\ \mathcal{I}_{\theta}(f^r;f^s)=\log\frac{p(f^r|f^s)}{p(f^r)},
\label{eq:mi_g2sd}
\end{split}
\end{equation}
which maximizes the mutual information between the regressor $f^r$ and the segmenter $f^s$.

\renewcommand{\tabcolsep}{5 pt}{
\begin{table*}[ht]
\footnotesize
\centering
\small
\resizebox{1\hsize}{!}{
  \begin{tabular}{ l | c | c  c | c  c | c  c | c  c | c  c }
  \toprule[1pt]
  \multirow{2}{*}{Method} & \multirow{2}{*}{Venue} & \multicolumn{2}{c|}{SHA} & \multicolumn{2}{c|}{SHB} & \multicolumn{2}{c|}{QNRF} & \multicolumn{2}{c|}{NWPU(V)} & \multicolumn{2}{c}{NWPU(T)}  \\
  {} & {} & MAE & RMSE & MAE & RMSE & MAE & RMSE & MAE & RMSE & MAE & RMSE  \\
  \hline
  GLoss~\cite{wan2021generalized} & CVPR'21 & 61.3 & 95.4 & 7.3 & 11.7 & 84.3 & 147.5  & - & - & 79.3 & 346.1 \\
  UEPNet~\cite{wang2021uniformity} & ICCV'21 & 54.6 & 91.2 & 6.4 & 10.9 & 81.1 & 131.7 & - & - & - & - \\
  P2PNet~\cite{song2021rethinking} & ICCV'21 & 52.8 & 85.1 & 6.3 & 9.9 & 85.3 & 154.5 & 77.4 & 362.0 & 83.3 & 553.9  \\
  DKPNet~\cite{chen2021variational} & ICCV'21 & 55.6 & 91.0 & 6.6 & 10.9 & 81.4 & 147.2 & 61.8 & 438.7 & 74.5 & 327.4 \\
  SASNeT~\cite{song2021sasnet} & AAAI'21 & 53.6 & 88.4 & 6.4 & 9.9 & 85.2 & 147.3 & - & - & - & - \\
  ChfL~\cite{shu2022crowd} & CVPR’22 & 57.5 & 94.3 & 6.9 & 11.0 & 80.3 & 137.6 & 76.8 & 343.0 & - & -  \\
  GauNet~\cite{cheng2022rethinking} & CVPR’22 & 54.8 & 89.1 & 6.2 & 9.9 & 81.6 & 153.7 & - & - & - & -  \\
  MAN~\cite{lin2022boosting} & CVPR'22 & 56.8 & 90.3 & - & - & 77.3 & 131.5 & 76.5 & 323.0 & - & -  \\
  S-DCNet (dcreg)~\cite{xiong2022discrete} & ECCV’22 & 59.8 & 100.0 & 6.8 & 11.5 & - & - & - & - & - & -  \\
  CLTR~\cite{liang2022end} & ECCV’22 & 56.9 & 95.2 & 6.5 & 10.6 & 85.8 & 141.3 & 61.9 & 246.3 & 74.3 & 333.8  \\
  DDC~\cite{ranasinghe2023diffuse} & CVPR’23 & 52.9 & 85.6 & \underline{6.1} & \underline{9.6} & \textbf{65.8} & \textbf{126.5} & - & - & - & -  \\
  PET~\cite{liu2023point} & ICCV’23 & \textbf{49.3} & \textbf{78.8} & 6.2 & 9.7 & 79.5 & 144.3 & 58.5 & \underline{238.0} & 74.4 & 328.5  \\
  STEERER~\cite{han2023steerer} & ICCV’23 & 54.5 & 86.9 & \textbf{5.8} & \textbf{8.5} & 74.3 & \underline{128.3} & \underline{54.3} & 238.3 & \underline{63.7} & 309.8  \\
  \hline
  mPrompt$_{\ddagger}$ (ours) & - & \underline{52.5} & 88.9 & \textbf{5.8} & \underline{9.6} & \underline{72.2} & 133.1 & \textbf{50.2} & \textbf{219.0} & \textbf{62.1} & \textbf{293.5}  \\ 
  mPrompt$_{\ddagger}${$^*$} (ours) & - & 53.2 & \underline{85.4} & 6.3 & 9.8 & 76.1 & 133.4 & 58.8 & 240.2 & 66.3 & \underline{308.4}  \\ 
  \bottomrule[1pt]
  \end{tabular}}
\caption{Performance comparisons. mPrompt$_{\ddagger}${$^*$} indicates that we extend the mPrompt to the pre-trained model (SAM-base). The best results are shown in \textbf{bold}, and the second-best results are \underline{underlined}.}
\label{table:state-of-the-art}
\end{table*}}

\section{Experiment}
\label{sec:Experiments}
\textbf{Dataset:} Experiments are carried out on four public crowd counting datasets including ShanghaiTechA/B~\cite{zhang2016single}, UCF-QNRF~\cite{idrees2018composition}, and NWPU~\cite{wang2020nwpu}.
\textbf{ShanghaiTech} includes PartA (\textbf{SHA}) and PartB (\textbf{SHB}), totaling $1,198$ images with $330,165$ annotated heads. SHA comprises $300$ training images and $182$ testing images with crowd sizes from $33$ to $3,139$. SHB includes $400$ training images and $316$ testing images with crowd sizes ranging from $9$ to $578$. The images are captured from Shanghai street views.
\textbf{UCF-QNRF} (\textbf{QNRF}) encompasses $1,535$ high-resolution images, $1.25$ million annotated heads with extreme crowd congestion, small head scales, and diverse perspectives. It is divided into $1,201$ training and $334$ testing images.
\textbf{NWPU} dataset comprises $5,109$ images, with $2,133,375$ annotated heads and head box annotations. The images are split to a training set of $3,109$ images, an evaluation set of $500$ images, and a testing set of $1,500$ images. NWPU(V) and NWPU(T) denote the validation and testing sets, respectively.

 {\textbf{Evaluation Metric:}} Mean Absolute Error (MAE) and Root Mean Squared Error (RMSE)~\cite{li2018csrnet,liu2019context} are used. They are defined as $M\!A\!E =  \frac{1}{N}\sum^{N}_{i=1}\vert \hat{C}_i - C_i \vert$ and $R\!M\!S\!E =  \sqrt{\frac{1}{N}\sum^{N}_{i=1}\vert \hat{C}_i - C_i \vert^{2}}$, where $N$ is the number of test images. $\hat{C}_{i}$ and ${C}_{i}$ respectively denote the estimated and ground truth counts of image $x_{i}$.

 {\textbf{Implementation Details:}} We resize images to a maximum length of $2,048$ pixels and a minimum of $416$ pixels, keeping the aspect ratio unchanged. Data augmentation includes random horizontal flipping, color jittering, and random cropping with a $400\times400$ pixel patch size. Ground-truth density maps are generated using a $15\times15$ Gaussian kernel. 
The network is trained using Adam~\cite{kingma2014adam} optimizer with learning rate of $1e^{-4}$. The batch size is $16$ and training on NWPU dataset takes about $25$ hours on four Nvidia V100 GPUs. Key parameters include $K=3$, $\lambda_{d}=1$, $\lambda_{s}=0.5$, $\lambda_{c}=0.5$ and $\kappa=50$. The network is constructed with the backbone HRNet-W40-C~\cite{wang2020deep} pretrained on ImageNet~\cite{krizhevsky2012imagenet} and random initialization of the remaining parameters. 
When adopting mPrompt for foundation models, we utilize SAM-base~\cite{kirillov2023segment}, chosen for its robust segmentation performance.
We train both networks for 700 epochs.

In Table~\ref{table:state-of-the-art}, the performance of mPrompt$_{\ddagger}$ is compared with state-of-the-art methods across four major datasets. mPrompt$_{\ddagger}$ consistently achieves impressive results in terms of MAE on all four datasets. mPrompt$_{\ddagger}$ consistently ranks within the top-2 for MAE performance across the datasets, highlighting the superior effectiveness of our model.

\subsection{Visualization Analysis}
\begin{figure*}[t!]
\centering
\begin{center}
    \includegraphics[width=1\textwidth]{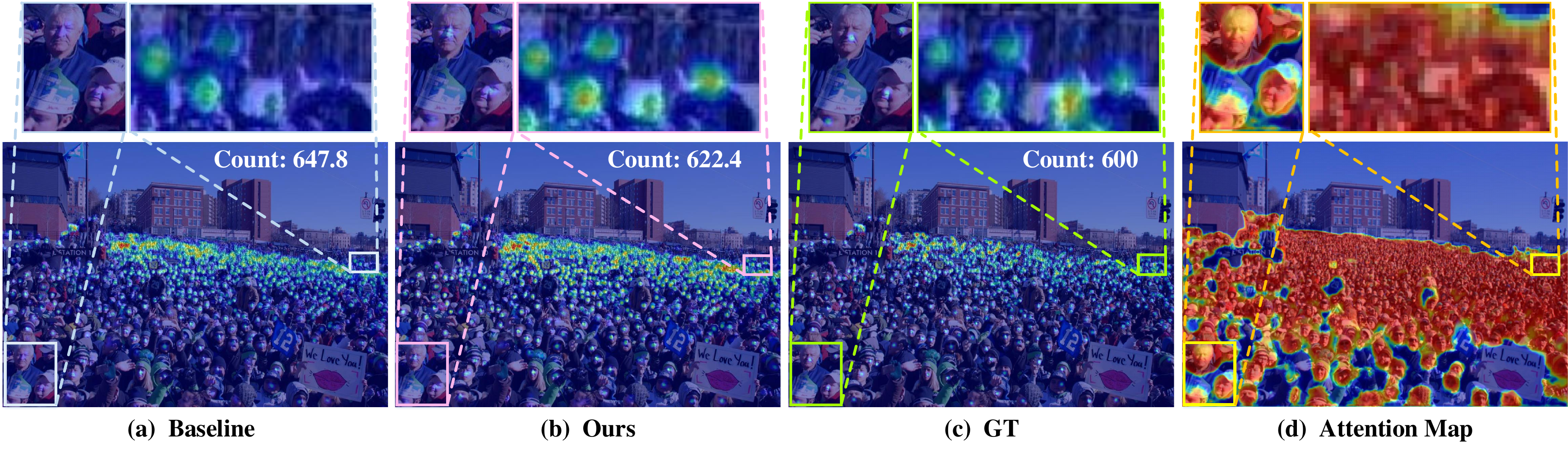}
\end{center}
\vspace{-0.5em}
\caption{Comparison of the density maps with/without context prompt. (Best viewed in color with zoom)}
\label{fig:density_map_visualization}
\end{figure*}

Fig.~\ref{fig:density_map_visualization} visualizes the predicted density maps and the attention map from a test image. mPrompt$_{\ddagger}$ generates more precise density maps compared with the baseline (mPrompt$_{rsg}$), at both dense and sparse regions. Particularly, after the context prompt learning, mPrompt$_{\ddagger}$ indeed isolates the accurate head regions as the regressor absorbs the context information from the segmenter.

\begin{figure*}[t!]
\centering
\begin{center}
    \includegraphics[width=1\textwidth]{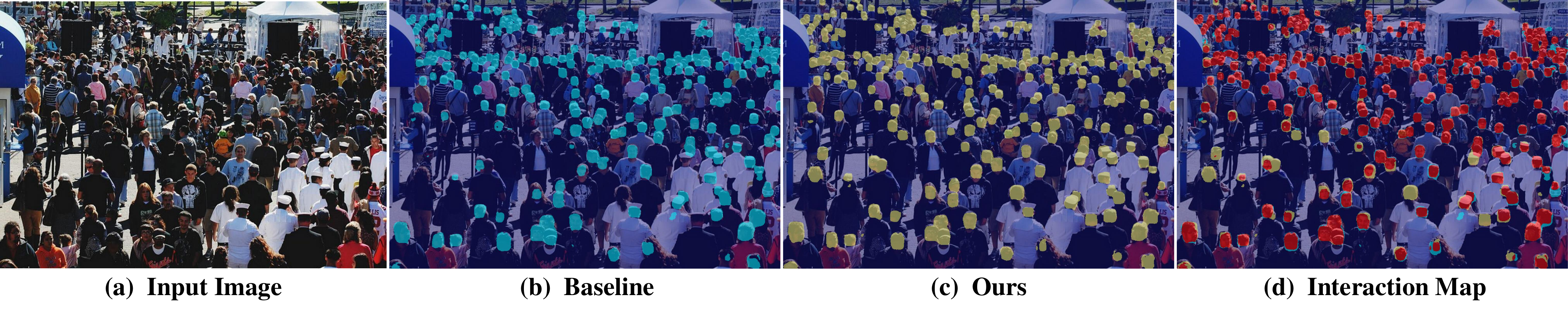}
\end{center}
\vspace{-0.5em}
\caption{Comparison of segmentation masks with/without point prompt. (Best viewed in color with zoom)}
\vspace{-0.5em}
\label{fig:segmentation_mask_visualization}
\end{figure*}

Fig.~\ref{fig:segmentation_mask_visualization} visualizes the segmentation maps predicted by mPrompt$_{rsg}$ and mPrompt$_{\ddagger}$. Specifically, we identify three types of regions when comparing these two segmentation maps.
Blue and yellow regions are generated by mPrompt$_{rsg}$ (baseline) and mPrompt$_{\ddagger}$, respectively.
Red regions represent the intersection of these two masks. 
One can see that mPrompt$_{\ddagger}$ improves head region segmentation by removing areas where the background is mistaken for a head and adding regions where the head is mistaken for background, compared to the baseline.
In order to evaluate the enhancement of the segmenter, we conduct an analysis of the Intersection over Union (IoU) between the head-box regions and the predicted mask on the NWPU dataset.
This investigation yields IoU scores of $46.5$ for mPrompt$_{\ddagger}$ and $38.7$ for mPrompt$_{rsg}$, respectively, thus providing quantitative evidence of the segmenter's improvement through mPrompt$_{\ddagger}$.
These validate the effect of point prompt learning, which finally contributes to the superior performance reported in Table~\ref{tab:performance_of_conponets}.

%
\subsection{Ablation Studies} 
\label{ablation}

%
\renewcommand{\tabcolsep}{12 pt}{
\begin{table*}[t!]
\small
	\begin{center}
 \resizebox{1\hsize}{!}{
		\begin{tabular}{l|c|c|cc|c|c|c|c|c}
			\toprule[1pt]
			\multirow{2}*{Methods} & \multirow{2}*{Regressor} & \multirow{2}*{Segmenter} & \multicolumn{2}{c|}{Point Prompt} & {Context} & \multirow{2}*{SHA} & \multirow{2}*{SHB} & \multirow{2}*{QNRF} & \multirow{2}*{NWPU(V)} \\
			{} & {} & {} & {Offline} & {Online} & {Prompt} & {} &
            {} & {} & {} \\
            \hline 
			mPrompt$_{reg}$ & $\checkmark$ & {} & {} & {} & {} & 59.4 & 7.8 & 85.5 & 65.7 \\
            \hline
            mPrompt$_{rsg}$ & $\checkmark$ & $\checkmark$ & {} & {} & {} & 58.4 & 7.1 & 83.2 & 64.3 \\
            \hline
    	  mPrompt$_{p\dagger}$ & $\checkmark$ & $\checkmark$ & $\checkmark$ & {} & {} & 54.8 & 6.2 & 78.9 & 59.2 \\
            \hline
    	  mPrompt$_{p\ddagger}$ & $\checkmark$ & $\checkmark$ & $\checkmark$ & $\checkmark$ & {} & 53.9 & 5.9 & 74.8 & 52.1 \\ 
            \hline
           mPrompt$_{c\dagger}$ & $\checkmark$ & $\checkmark$ & {} & {} & $\checkmark$ & 55.3 & 6.4 & 79.4 & 62.0 \\ 
           \hline
		  mPrompt$_{\dagger}$ & $\checkmark$ & $\checkmark$ & $\checkmark$ & {} & $\checkmark$ & 54.1 & 6.1 & 76.7 & 56.5 \\
           \hline
		  mPrompt$_{\ddagger}$ & $\checkmark$ & $\checkmark$ & $\checkmark$ & $\checkmark$ & $\checkmark$ & \textbf{52.5} & \textbf{5.8} & \textbf{72.2} & \textbf{50.2} \\
			\bottomrule[1pt]
		\end{tabular}
  }
	\end{center}
 \vspace{-0.5em}
\caption{Ablation study of mPrompt components about MAE.}
\vspace{-0.5em}
\label{tab:performance_of_conponets}
\end{table*}}
\vspace{-0.5em}
\renewcommand{\tabcolsep}{12 pt}{
\begin{table*}[ht]
\footnotesize
\centering
\small
\resizebox{1\hsize}{!}{
  \begin{tabular}{ c | c | c | c  c | c  c | c  c | c  c }
  \toprule[1pt]
  \multirow{2}{*}{Backbones} & \multirow{2}{*}{\#Params(M)} & \multirow{2}{*}{GFLOPs} & \multicolumn{2}{c|}{mPrompt$_{reg}$} & \multicolumn{2}{c|}{mPrompt$_{rsg}$} & \multicolumn{2}{c|}{mPrompt$_{\dagger}$} & \multicolumn{2}{c}{mPrompt$_{\ddagger}$}  \\
  {} & {} & {} & MAE & RMSE & MAE & RMSE & MAE & RMSE & MAE & RMSE   \\
  \midrule      
  \multicolumn{11}{c}{\textbf{CNN architecture}}  \\
  \midrule 
  VGG19~\cite{simonyan2014very} & 12.6 & 19.3 & 64.0 & 112.5  & 62.6 & 106.5 & 61.4 & 100.9 & 60.9 & 106.1  \\ 
  HRNet~\cite{wang2020deep} & 33.1 & 62.1 & 59.4 & 96.7 & 58.4 & 95.8 & 54.1 & 92.8 & 52.5 & 88.9  \\
  \midrule      
  \multicolumn{11}{c}{\textbf{Transformer architecture}}  \\
  \midrule 
  Swin~\cite{liu2021Swin} & 7.4 & 11.6 & 63.9 & 105.5 & 61.8 & 100.0 & 61.1 & 99.3 & 59.3 & 98.8  \\
  SAM~\cite{kirillov2023segany} & 7.7 & 13.5 & 60.4 & 98.3 & 59.5 & 98.8 & 55.2 & 89.5 & 53.2 & 85.4  \\ 
  \bottomrule[1pt]
  \end{tabular}}
  \vspace{-0.5em}
\caption{Comparison of backbones on the SHA dataset is paired with an analysis of learnable parameters and FLOPs for a standard input size of ($3\times 224\times 224$) when training.}
\vspace{-0.5em}
\label{tab:robustness_of_backbone}
\end{table*}}

\renewcommand{\tabcolsep}{6 pt}{
\begin{table}[ht]
\small
\begin{center}
\resizebox{1\hsize}{!}{
  \begin{tabular}{l|cc|c|c|c|c}
  \toprule[1pt]
  \multicolumn{7}{c}{\textbf{No Mutual Prompt Learning}}  \\
  \hline
  \multirow{2}*{Method} & \multicolumn{2}{c|}{Seg label} & \multirow{2}*{SHA} & \multirow{2}*{SHB} & \multirow{2}*{QNRF} & \multirow{2}*{NU(V)} \\		
  {} & {point} & {box} & {} & {} & {} & {} \\
  \hline 
  mPt$_{rsg*}$ & $\checkmark$ & {} & 58.8  & 7.5 & 84.3 & 66.8 \\
  \hline
  mPt$_{rsg}$ & {} & $\checkmark$ &  58.4 & 7.1 & 83.2 & 64.3 \\
  \hline
  \multicolumn{7}{c}{\textbf{Mutual Prompt Learning}}  \\
  \hline
  \multirow{2}*{Method} & \multicolumn{2}{c|}{$\boldsymbol{m_p}$} & \multirow{2}*{SHA} & \multirow{2}*{SHB} & \multirow{2}*{QNRF} & \multirow{2}*{NU(V)} \\
  {} & {mPt$_{rsg*}$} & {mPt$_{rsg}$} & {} & {} & {} & {} \\
  \hline
  mPt$_{\ddagger\emptyset}$ & \multicolumn{2}{c|}{$\emptyset$} &  54.6 & 6.3 & 73.9 & 52.1 \\
  \hline
  mPt$_{\ddagger*}$ & $\checkmark$ & {} &  54.3 & 6.4 & 73.4 & 52.7 \\
  \hline
  mPt$_{\ddagger}$ & {} & $\checkmark$ &  52.5 & 5.8 & 72.2 & 50.2 \\
  \bottomrule[1pt]
  \end{tabular}
}
\end{center}
\caption{Performance when adopting different pseudo masks. Due to space constraints, we use the abbreviations NU(V) and mPt to respectively refer to NWPU(V) and mPrompt.}
\vspace{-0.5em}
\label{tab:performance_of_mask}
\end{table}}

%
{\textbf{No Prompt.}} 
The baseline mPrompt$_{reg}$ consists only a regressor.
By introducing the segmenter and employing pseudo mask as supervision, mPrompt$_{reg}$ develops to mPrompt$_{rsg}$. 
In Table~\ref{tab:performance_of_conponets}, mPrompt$_{reg}$ harnesses the robust features of HRNet (truncated at \textit{stage4}), achieving competitive MAE performances of $59.4$, $7.8$, $85.5$, and $65.7$ on SHA, SHB, QNRF, and NWPU(V) datasets, respectively.
mPrompt$_{rsg}$ surpasses mPrompt$_{reg}$, highlighting the significance of introducing the segmenter and signifying the effective utilization of spatial head information.

{\textbf{Point Prompt.}}
With offline and online point prompt, mPrompt$_{rsg}$ promotes to mPrompt$_{p\dagger}$ and mPrompt$_{p\ddagger}$, respectively.
mPrompt$_{p\dagger}$ achieves better performance, reaching MAEs of $54.8$, $6.2$, $78.9$, and $59.2$ on SHA, SHB, QNRF, and NWPU(V) datasets, respectively. 
mPrompt$_{p\ddagger}$ further reduces the MAEs to $53.9$, $5.9$, $74.8$, and $52.1$ on these four datasets.

{\textbf{Context Prompt.}}
In Table~\ref{tab:performance_of_conponets}, when adopting $\mathcal{L}_{con}$ to mPrompt$_{rsg}$, our mPrompt$_{c\dagger}$ delivers 
a performance gain on these datasets, indicating the necessity of spatial information for regressing implemented in this explicit manner.

 {\textbf{Mutual Prompt.}}
In Table~\ref{tab:performance_of_conponets}, both mPrompt${\dagger}$ and mPrompt${\ddagger}$ achieves satisfying performances, and our final variant mPrompt${\ddagger}$ delivers MAEs of $52.5$, $5.8$, $72.2$, and $50.2$ on SHA, SHB, QNRF, and NWPU(V) datasets, respectively. 
Comparing with mPrompt$_{reg}$, a significant performance gain is achieved, reducing MAE by $6.9$, $2.0$, $13.3$, and $15.5$, respectively. 
These ablation studies validate the efficacy of the components of mPrompt. 

\textbf{Pseudo masks.}
\label{sec:point_mask}
We use a segmenter pretrained on NWPU box annotations to obtain the offline pseudo mask $\boldsymbol{m}_{p}$.
A natural question arises: \textit{Can we generate $\boldsymbol{m}_{p}$ using a segmenter pretrained only with the point annotations, or even directly set $\boldsymbol{m}_{p}$ as $\emptyset$?}
To explore this, we pretrain mPrompt$_{rsg*}$ using only the point annotations of the corresponding dataset to generate the segmentation masks (\textit{i.e.}, point-based pseudo mask).
In Table~\ref{tab:performance_of_mask}, mPrompt$_{rsg*}$ underperforms mPrompt$_{rsg}$ due to the inaccuracy of the segmentation label.
By setting $\boldsymbol{m_{p}}$ to $\emptyset$ and utilizing pseudo masks generated from mPrompt$_{rsg*}$ and mPrompt$_{rsg}$ in mutual prompt learning, we obtain mPrompt$_{\ddagger\emptyset}$, mPrompt$_{\ddagger*}$ and mPrompt$_{\ddagger}$, respectively. 
mPrompt$_{\ddagger*}$ performs similarly to mPrompt$_{\ddagger\emptyset}$, as $\boldsymbol{m}_{p}$ indeed introduces no extra spatial information when only utilizing the pseudo masks generated from mPrompt$_{\ddagger*}$.
Even with $\boldsymbol{m}_{p}$ set to $\emptyset$, mPrompt$_{\ddagger\emptyset}$ still significantly outperforms mPrompt$_{reg*}$, highlighting the effectiveness of mutual prompt learning.

\textbf{Backbone Architectures.}
We replace HRNet-W40-C with other commonly-used backbones (VGG19~\cite{simonyan2014very}, Swin~\cite{liu2021Swin} and SAM~\cite{kirillov2023segany}).
Table~\ref{tab:robustness_of_backbone} reveals that mPrompt$_{\ddagger}$ continues to outperform mPrompt$_{reg}$, mPrompt$_{rsg}$, and mPrompt$_{\dagger}$, achieving significant MAE reductions. 
Furthermore, we have extended the mPrompt to foundational models, such as the SAM~\cite{kirillov2023segany} and Swin~\cite{liu2021Swin}. As shown in Table~\ref{tab:robustness_of_backbone}, mPrompt$_{\ddagger}$ (SAM based) shows performance marginally below mPrompt$_{\ddagger}$, yet with only about $\frac{1}{4}$ the training parameters and $\frac{1}{5}$ the FLOPs of the latter.
For crowd counting, given the backbone is static and only the prompt module is learnable, the Swin Transformer, pretrained for classification, underperforms compared to SAM~\cite{kirillov2023segany}.
This mainly attributes to Swin's representational knowledge is less aligned with crowd counting comparing with SAM.

\begin{figure}[t]
    \centering
    \includegraphics[width=0.70\linewidth]{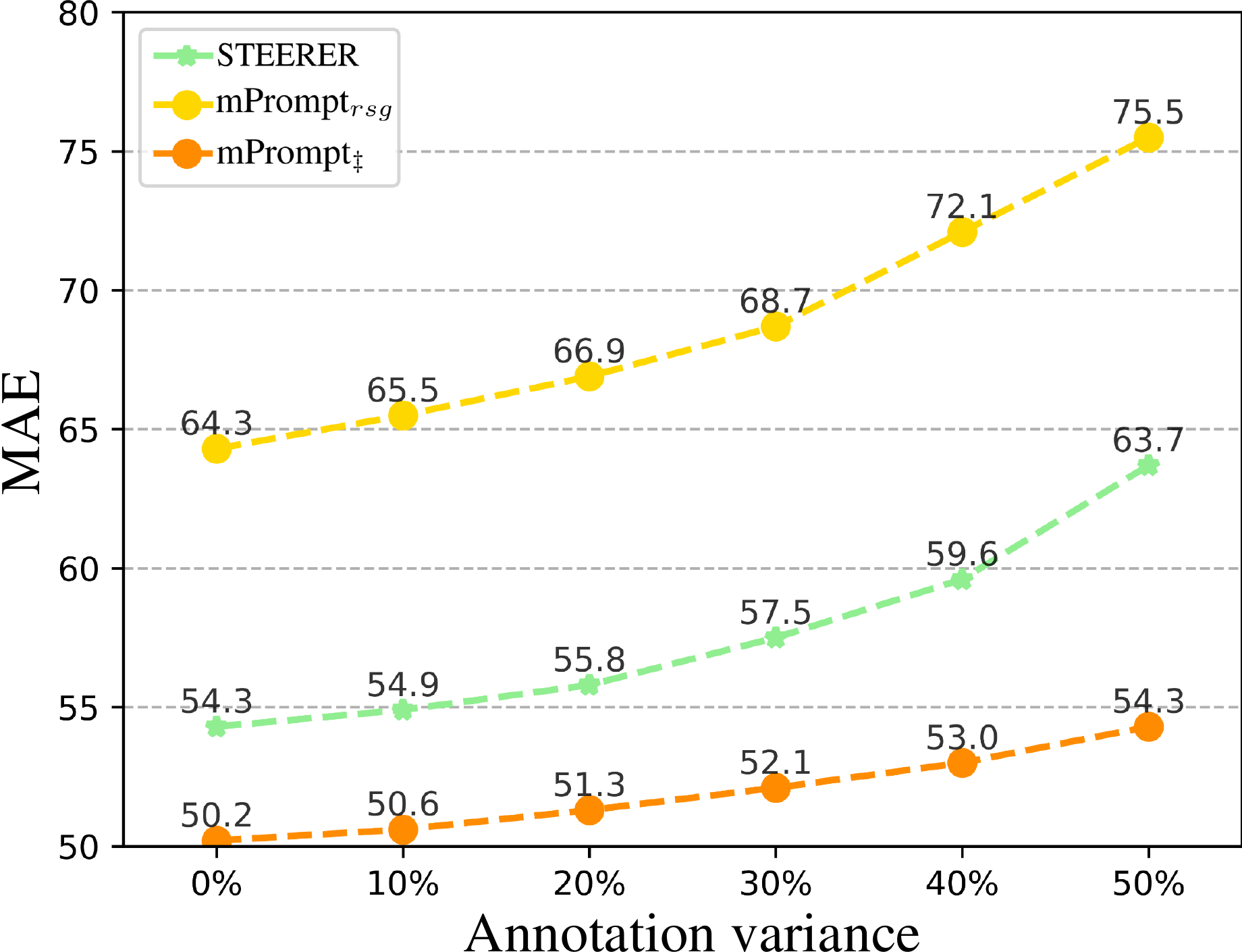}
    \vspace{-0.5em}
    \caption{Performance on NWPU when training with different annotation variance.}
    \label{fig:noise}
\end{figure}

\textbf{Robustness to Annotation Variance.}
\label{sec:point_noise}
To assess the robustness of mutual prompt learning against box annotation variance, we conduct an experiment on NWPU, observing performance changes with varying box annotations.
Specifically, we add uniform random noise, ranging from $0$ to $50\%$, of the box height, to the official annotated boxes.
Fig.~\ref{fig:noise} reveals that our mPrompt$_\ddagger$ is only mildly affected by different noise levels, while mPrompt$_{rsg}$ and STEERER~\cite{han2023steerer} suffer from severe performance degradation.
This demonstrates the robustness of our mPrompt$_\ddagger$ to annotation variance.

\section{Conclusions}
\vspace{-0.5em}
We proposed a mutual prompt learning approach, to enhance context information while mitigating the impact of point annotation variance in crowd counting.
mPrompt incorporates a shared backbone, a density map regressor for counting, a head segmenter for foreground and background distinction.
The mutual prompt learning strategy maximized the mutual information gain of the segmenter and regressor. 
Experimental results on four public datasets affirm the efficacy and superiority of our method.
While we primarily focus on crowd density maps in this study, mPrompt has potential applications in areas with scarce or noisy labeling information, such as crowd localization, object detection, and visual tracking. We aim to explore these applications in the future work.

{
    \small
    \bibliographystyle{ieeenat_fullname}
    \bibliography{main}
}

\newpage
\appendix

\section{Details of Training a Segmenter}
\noindent\textbf{Training via point Annotation.}
For methods~\cite{zhao2019leveraging, shi2019counting, modolo2021understanding}, they adopt point-segmentation map $P$ as the target of the segmenter, as formulated in Equation~\eqref{eqn:dot_seg_loss}.
\begin{equation}
\label{eqn:dot_seg_loss}
   \mathcal{L}_{poi} = -\sum^{N}_{i=1}\sum^{(H,W)}_{(h=1,w=1)}p(h, w)\log \hat{p}(h, w),
\end{equation}
where $N$ is the batch size, $H$ and $W$ are the height and width of image $x_i$, respectively. $p(h, w)$ represents the value of position $(h, w)$ in binarized ground-truth density map $P_{i}$, and $\hat{p}(h, w)$ is the corresponding predicted value.
To this end, we build the mPrompt$_{poi}$ under the loss $\mathcal{L}_{poi}$, formulated as
\begin{equation}
\label{eqn:base}
\boldsymbol{\mathcal{L}} =  \mathcal{L}_{den}+\lambda_{s}\mathcal{L}_{poi},
\end{equation}
where $\lambda_{s}$ is a super-parameter to balance the two losses.

As elucidated in scratch, mPrompt$_{poi}$ exhibits a challenge in assimilating spatial information. This limitation primarily stems from the fact that the targets for both the segmenter and regressor are manually created from dot annotations, which intrinsically do not convey any spatial information.

\noindent\textbf{Training via Box Annotation.}
To strengthen the segmenter's ability in integrating spatial information, we pretrain it using head-box annotations of NWPU~\cite{wang2020nwpu} dataset, and generate pseudo mask ($m_p$) of all datastes for mutual prompt learning.
Concretely, suppose $B_{i, v}$ is the $v$-th box annotation in the image $x_i$ and its annotation is $(x_l, y_l, x_r, y_r)$ representing upper-left corner and lower-right corner. 
The head box region $R_{i, v}$ is defined as:
\begin{equation} \label{eqn:box_region}
  \begin{split}
  &B^{i,v}_{x_{min}} = x_l,  B^{i,v}_{y_{min}} = y_l, 
  B^{i,v}_{x_{max}} = x_r,  B^{i,v}_{y_{max}} = y_r \\
  &R^{i}_{v} = \left\{(x, y)|B^{i,v}_{x_{min}} \leq x \leq B^{i,v}_{x_{max}}, B^{i,v}_{y_{min}} \leq y \leq B^{i,v}_{y_{max}} \right\} \\
  \end{split}
\end{equation}

Then the ground-truth box-segmentation map $S_{i}$ for pretraining the head segmenter is defined as
\begin{equation}
\label{eqn:box_seg}
   S_{i} = \cup^{V_i}_{v=1}R^{i}_{v},
\end{equation}
where $V_i$ denotes the number of boxes in image $x_i$ and $\cup$ indicates the union operator.
In this case, for $s(h,w)$ which indicates the value of position $(h,w)$ in the $S_{i}$ , we have
\begin{equation}
\label{seg_ind}
    s(h,w)=\mathbb{I}\left((h,w)\in S_{i}\right).
\end{equation}
The function $\mathbb{I}(cond)$ is the indicator function, which is equal to $1$ only if the condition holds, and $0$ otherwise.
We utilize $\mathcal{L}_{box}$ to encourage the segmenter to predict a value of $1$ for positions falling within any heads, and a value of $0$ for positions outside them. Formulately, $\mathcal{L}_{box}$ is defined as
\begin{equation}
\label{eqn:bce}
   \mathcal{L}_{box} = -\sum^{N}_{i=1}\sum^{(H,W)}_{(h=1,w=1)}s(h, w)\log \hat{s}(h, w),
\end{equation}
where $N$ is the batch size, $H$ and $W$ are the height and width of image $x_i$, resp.
$s(h, w)$ represents the value of position $(h, w)$ in $S_{i}$, and $\hat{s}(h,w)$ is the corresponding predicted value. 

Similar to $\mathcal{L}_{poi}$ and $\mathcal{L}_{box}$, $\mathcal{L}_{seg}$ in manuscript is implemented on density map ($\hat{y}$) and the pseudo mask ($m_p$) generated by the pretrained segmenter.
Finally, mPrompt is trained with $\mathcal{L}_{seg}$ as follows:
\begin{equation}
\label{eqn:box_base}
{\mathcal{L}}=\mathcal{L}_{den}+\lambda_{s}\mathcal{L}_{seg}
\end{equation}
where $\lambda_{s}$ balances the two losses.

\section{Details of Extension to Foundation Model}
The broadly acknowledged foundational model SAM~\cite{kirillov2023segany} for image segmentation functions at the pixel level, similar to crowd counting tasks based on density map method. Therefore, SAM has been selected as the foundation model for extending our mPrompt approach, aiming at modifying the hidden representations of a frozen pre-trained model.

\noindent\textbf{The position of adapter.}
The pre-trained SAM's image encoder, equipped with adapter modules identical to the scaled parallel adapter~\cite{he2021towards}, has supplanted the backbone of our previous architecture. We fixed the parameters of the image encoder, making only a few parameters trainable, including the adapter modules, regressor and segmenter. 
Specifically, the image encoder is composed of 12 stacked blocks, each containing two types of sublayers: multi-head self-attention (MHA) and a fully connected feed-forward network (FFN). 

Adapters are utilized to modify the outputs of MHA and FFN in the transformer blocks. The output from the last adapter module serves as the input for the segmenter. Throughout this process, the adapter modules function as a context prompt (akin to mask prompts in SAM), referred to here as learnable prompt modules. 

\noindent\textbf{The performance of training with adapter.}
To further validate the potential performance enhancement of mPrompt on foundation model, we evaluated its effectiveness on SHA under the same architecture (image encoder + adapter + regressor + segmenter), but with varied training strategies. These strategies include full fine-tuning without mutual prompt learning, adapter training without mutual prompt learning, and learnable prompt modules with mutual prompt learning. As presented in Table~\ref{tab:finetune_adapter}, it is evident that our method offers significant improvement opportunities when applied to foundational model.

\renewcommand{\tabcolsep}{11 pt}{
\begin{table}[ht]
\small
\begin{center}
\resizebox{0.7\hsize}{!}{
  \begin{tabular}{c c c}
  full ft & adapter & mPrompt \\
  \hline 
  {54.8} & {56.2} & {53.2} \\
\end{tabular}
}
\end{center}
\caption{Performance on SHA about MAE, when adopting different training strategies.}
\label{tab:finetune_adapter}
\end{table}}

%
\section{Analysis of Convergence Speed via Context Constraint}

\begin{figure}[ht]
\centering

\begin{overpic}[scale=0.28]{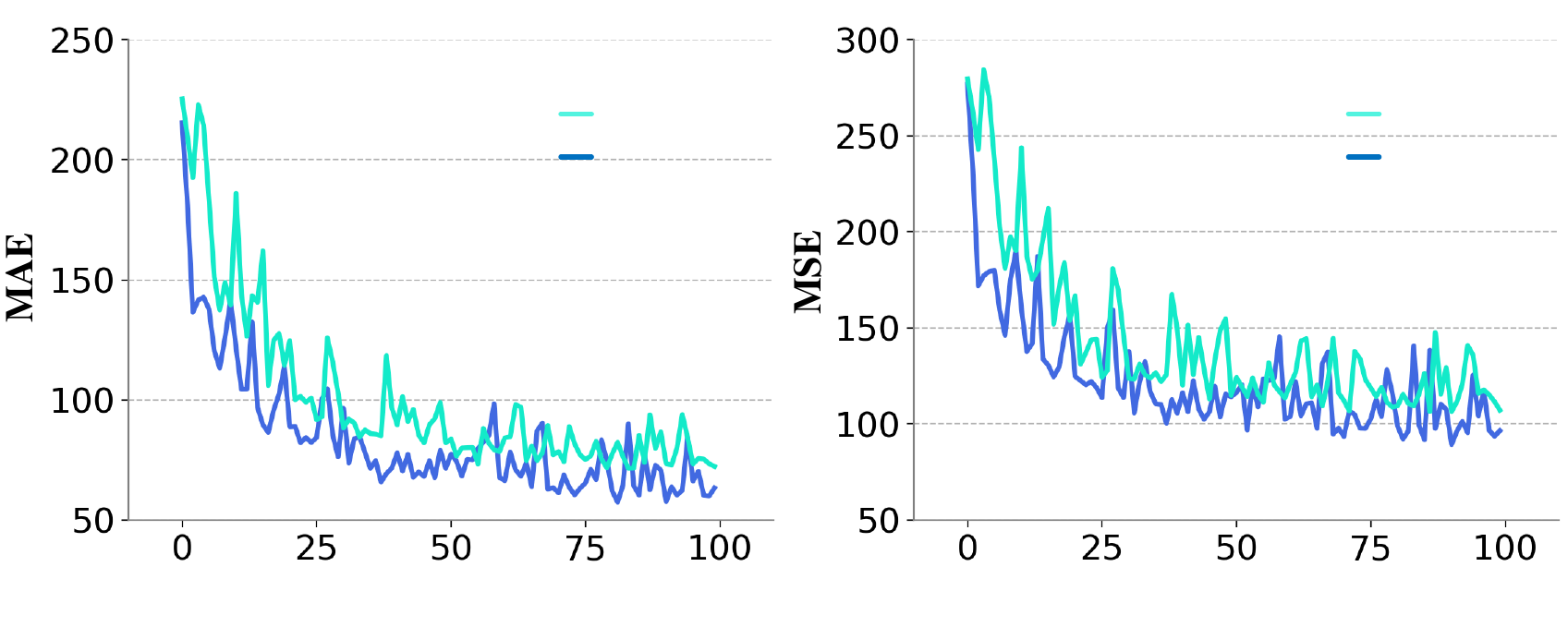}

\put(26, 1){\scalebox{0.9}{Epoch}}
\put(76, 1){\scalebox{0.9}{Epoch}}

\put(39, 29.5){\scalebox{0.7}{$\lambda_{con}$= 1}}
\put(89, 29.5){\scalebox{0.7}{$\lambda_{con}$= 1}}

\put(39, 32.5){\scalebox{0.7}{$\lambda_{con}$= 0}}
\put(89, 32.5){\scalebox{0.7}{$\lambda_{con}$= 0}}

\end{overpic}
\scriptsize

\caption{MAEs and MSEs of $\lambda_{con}=0$ and $\lambda_{con}=1$.}\label{fig:mae_mse_convergence}

\end{figure}
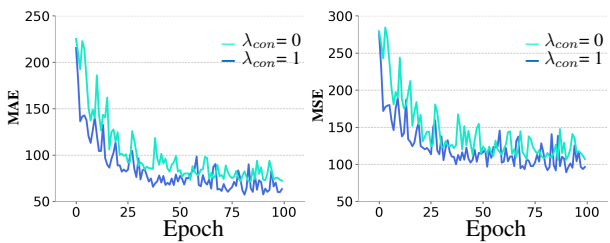

We have introduced $\mathcal{L}_{con}$ as a mechanism to guide non-zero values of the density map ($\hat{y}$) to fall within mask ($\hat{m}$), a crucial factor for achieving rapid convergence of the regressor.
To validate this assertion, we conducted an experiment comparing convergence speed based on the inclusion or exclusion of $\mathcal{L}_{con}$.
Figure~\ref{fig:mae_mse_convergence} displays the Mean Absolute Errors (MAEs) and Mean Squared Errors (MSEs) of the initial 100 epochs during the training process on SHA.
The green curve represents the model trained without $\mathcal{L}_{con}$ ($\lambda_{con}=0$), while the blue curve signifies the model trained with $\lambda_{con}=1$.
Upon examining these results, it becomes clear that both the MAEs and MSEs of the model trained with $\lambda_{con}=1$ are consistently lower than those of the model trained without it.
These findings underscore that the incorporation of $\lambda_{con}$ effectively aids in achieving faster convergence of the regressor.

\section{Hyper-parameters}
\label{sec:hyper_parameters}
\begin{figure*}
    
    \begin{minipage}[c]{0.55\textwidth}
    \centering
    \begin{overpic}[scale=0.35]{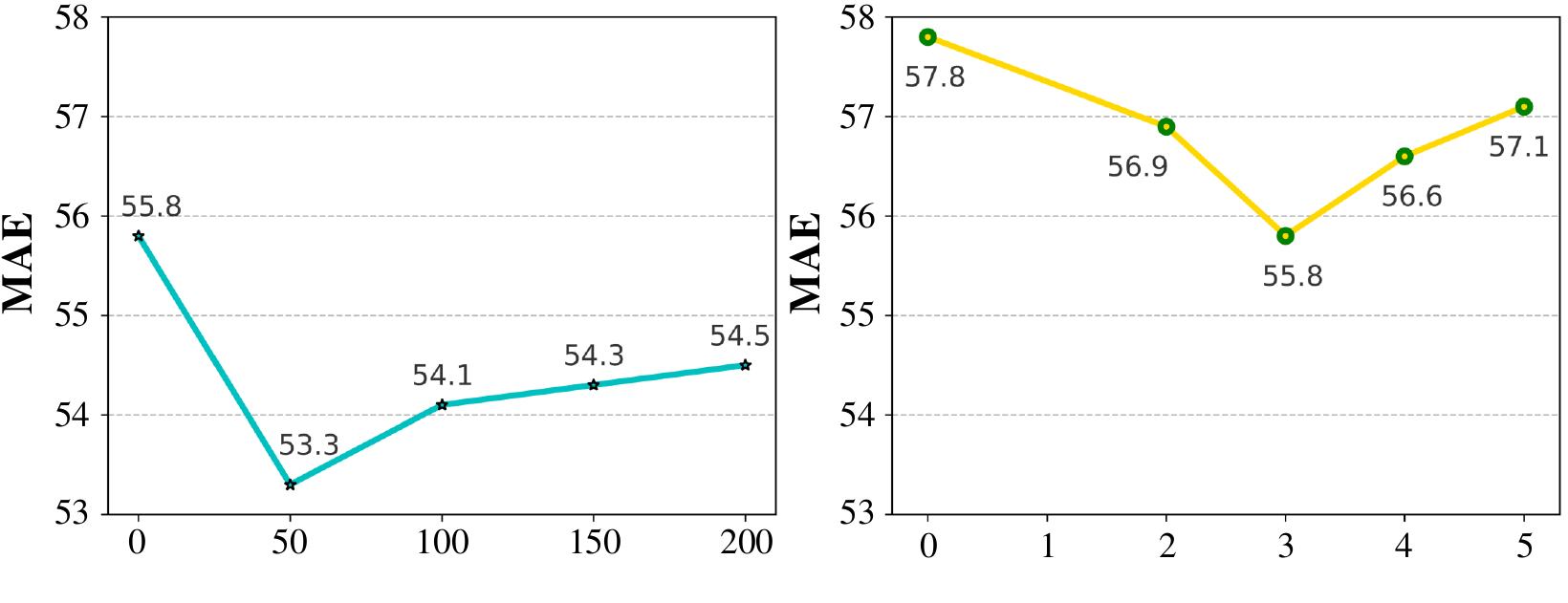}
    \put(27, 0.0){\scalebox{0.8}{$K$}}
    \put(74, 0.0){\scalebox{0.9}{$\kappa$}}
    \end{overpic}
    \figcaption{Evaluation of $K$ and $\kappa$ on SHA.}
    \label{fig:k_e}
    \end{minipage}
    \hspace{8mm}
    \begin{minipage}[c]{0.28\textwidth}
    \centering
    \small
    \begin{tabular}{p{0.6cm}|p{0.6cm}|p{0.6cm}|p{0.6cm}}
    \toprule[1.5pt]
    \multicolumn{1}{p{0.6cm}|}{\multirow{2}*{$\lambda_{d}$}} &  \multicolumn{1}{p{0.6cm}|}{\multirow{2}*{$\lambda_{s}$}} &  \multicolumn{1}{p{0.6cm}|}{\multirow{2}*{$\lambda_{c}$}} &
    {\multirow{2}*{MAE}} \\
    \multicolumn{1}{p{0.6cm}|}{} & \multicolumn{1}{p{0.6cm}|}{} &
    \multicolumn{1}{p{0.6cm}|}{} \\
 
    \hline
    1 & 0 & 0 & 58.4  \\
    1 & 1 & 0 & 55.9  \\
    1 & 1 & 1 & 53.3  \\
    1 & 0.5 & 1 & 53.7  \\
    1 & 1 & 0.5 & 53.5 \\
    \textbf{1} & \textbf{0.5} & \textbf{0.5} & \textbf{52.5}  \\
    \bottomrule[1.5pt]
    \end{tabular}
    \tabcaption{Regularization factors.}
    \label{table:loss_weight}    
    \end{minipage}
\end{figure*}

We investigate hyper-parameters including $K$ in $K$-NN, epoch $\kappa$ to begin online point prompt, and loss weights $\lambda_{d}$, $\lambda_{s}$, $\lambda_{c}$. Grid-search is infeasible due to computational constraints. 
Initially, we explore $K$ while fixing other settings at $\kappa=0$, $\lambda_{d}=\lambda_{s}=\lambda_{c}=1$. Fig.~\ref{fig:k_e} reveals $K=3$ as optimal, yielding the lowest MAE of $55.0$. Using $K=3$, we find $\kappa=50$ reduces MAE to $53.3$.
Two conclusions can be drawn: 1) An appropriate $K$ can reduce MAE by approximately a gap of $1$. For very large $K$ values ($e.g.$, $K=5$), the performance is similar to $K=0$. This occurs because a large $K$ means $m_{K}$ covers nearly the entire image, rendering $m_{K}$ almost ineffective. 2) $\kappa=50$ delivers the best MAE, indicating the learning of regressor is fast due to the deployment of $\mathcal{L}_{con}$.

In Table~\ref{table:loss_weight}, our approach, even when applied with a basic hyper-parameter search, successfully reduces MAE to $52.5$. This is achieved with the settings $\lambda_{d}=1$, $\lambda_{s}=0.5$, and $\lambda_{c}=0.5$. 
We further have the following two conclusions: 
1) Employing only $\mathcal{\lambda}_{d}$, the network still reduces MAE to MAE $58.4$, a marginally performance gain compared to mPrompt$_{reg}$. This affirms the rationality of incorporating the attention as implicit spatial context from segmenter to regressor. Additionally, when $\mathcal{\lambda}_{s}$ and $\mathcal{\lambda}_{c}$ are incorporated into the network learning process, we observe further performance gain, underlining the effect of these elements.
2) Upon examining the last three row in Table~\ref{table:loss_weight}, we confirm that appropriate selection of loss weights can further help enhance performance. 
%

\section{Visualization of mPrompt on Tackling Point Annotation Variance in Highly Congested Scenarios}
\begin{figure*}[t]
    \centering
    \includegraphics[width=1\textwidth]{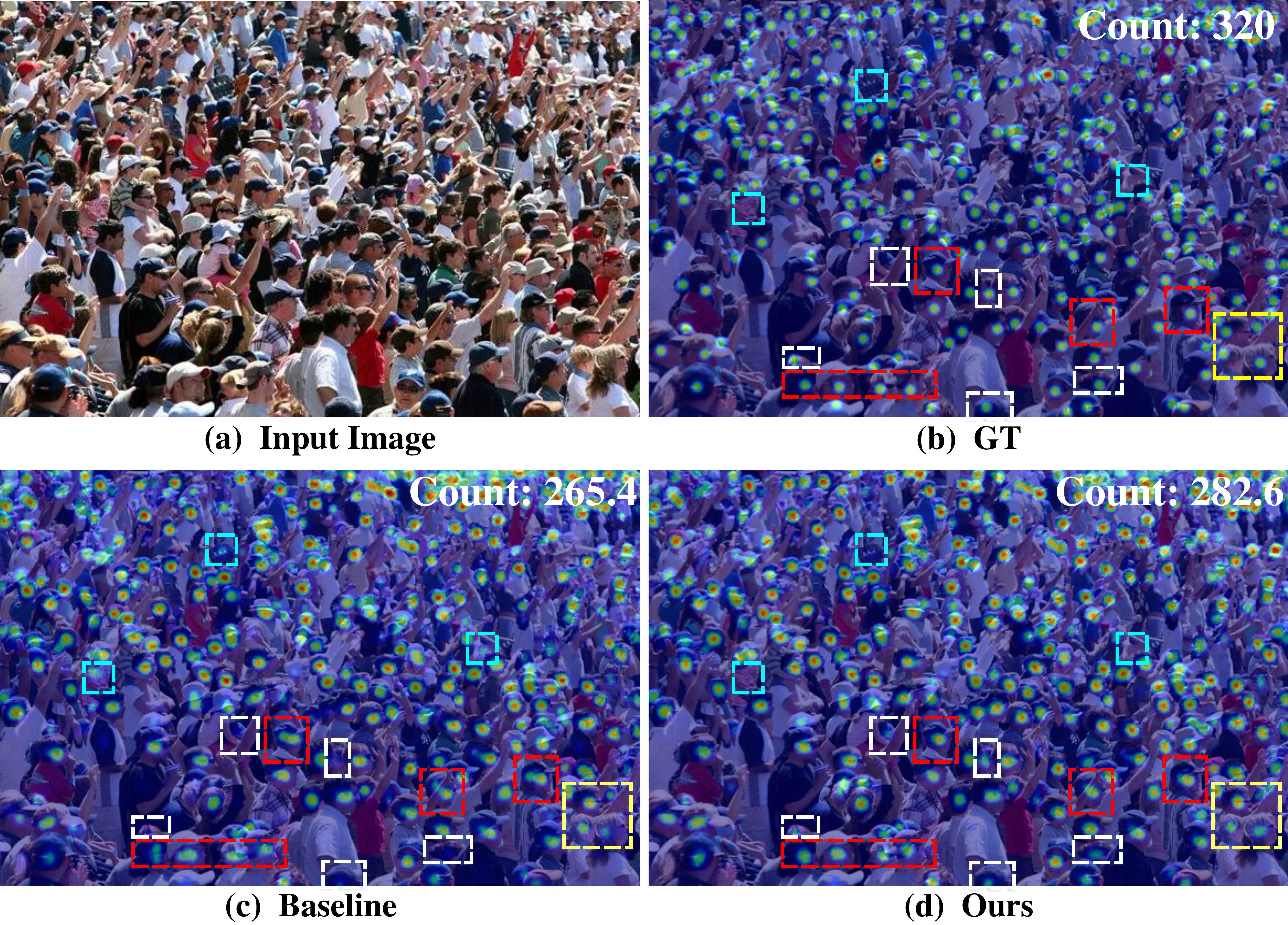}
    \caption{Visualization of density maps. (Best viewed in color)}
    \label{fig:supp_point_bias}
\end{figure*}

In this section, we delve deeper into the validation of mPrompt's efficiency in addressing point annotation variance in highly crowded scenarios.
Figure~\ref{fig:supp_point_bias} exhibits the respective density maps as predicted by mPrompt$_{\ddagger}$ and mPrompt$_{reg}$ (named ``baseline'' in the image). In the presented graphic, we have highlighted certain regions using color-coded boxes for ease of understanding.
The areas shaded in blue represent the background regions, wherein mPrompt$_{reg}$ exhibits high activations, contrasting with mPrompt$_{\ddagger}$, which does not.
In the regions designated by red boxes, we demonstrate the head areas where mPrompt$_{reg}$ displays inaccurate density blobs, whereas mPrompt$_{\ddagger}$ successfully predicts accurate blobs.
The white boxes highlight the head areas that mPrompt$_{reg}$ failed to identify correctly, while, conversely, mPrompt$_{\ddagger}$ delivers correct activations.
Lastly, the yellow boxes underscore the head regions where mPrompt$_{reg}$ exhibits activations displaced from the center of the corresponding boxes. In contrast, mPrompt$_{\ddagger}$ generates density blobs precisely at the center of the heads.
In summary, for all these four identified situations, mPrompt$_{\ddagger}$ consistently outperforms mPrompt$_{reg}$ in accurately predicting head density blobs.

\section{Visualization of mPrompt on Predicting Density Maps}
\begin{figure*}[ht]
    \centering
    \includegraphics[width=1\textwidth]{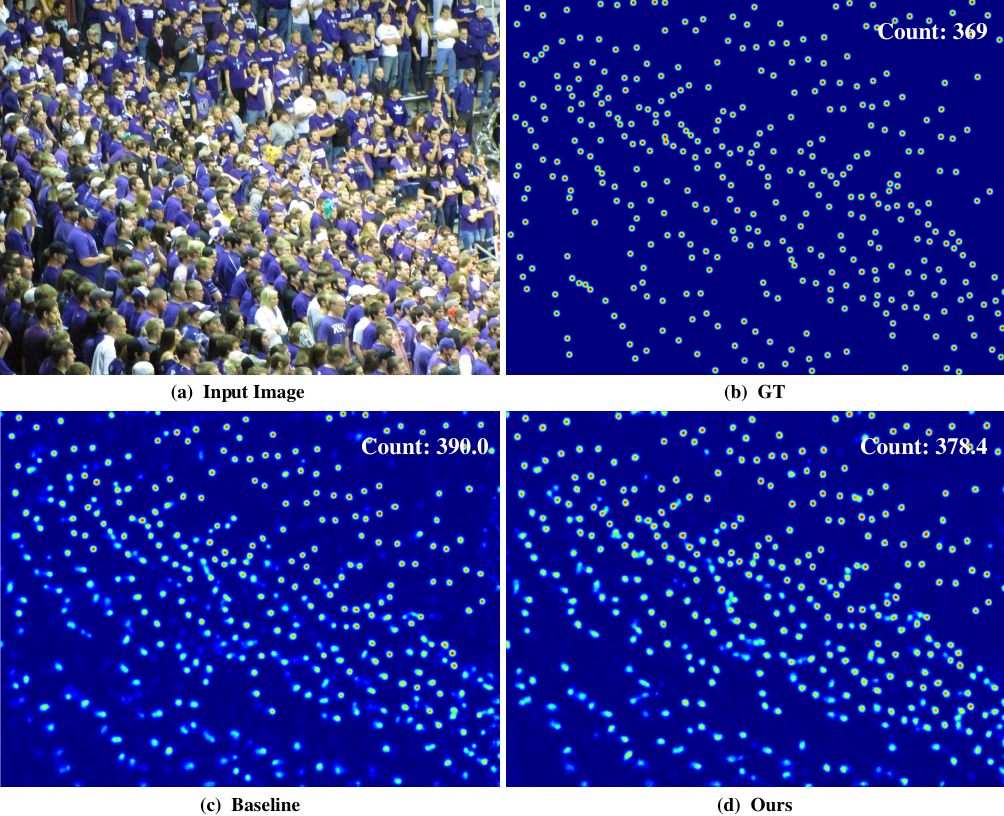}
    \caption{Visualization of predicted density maps from SHA.}
    \label{fig:density_map_sha}
\end{figure*}

\begin{figure*}[ht]
    \centering
    \includegraphics[width=1\textwidth]{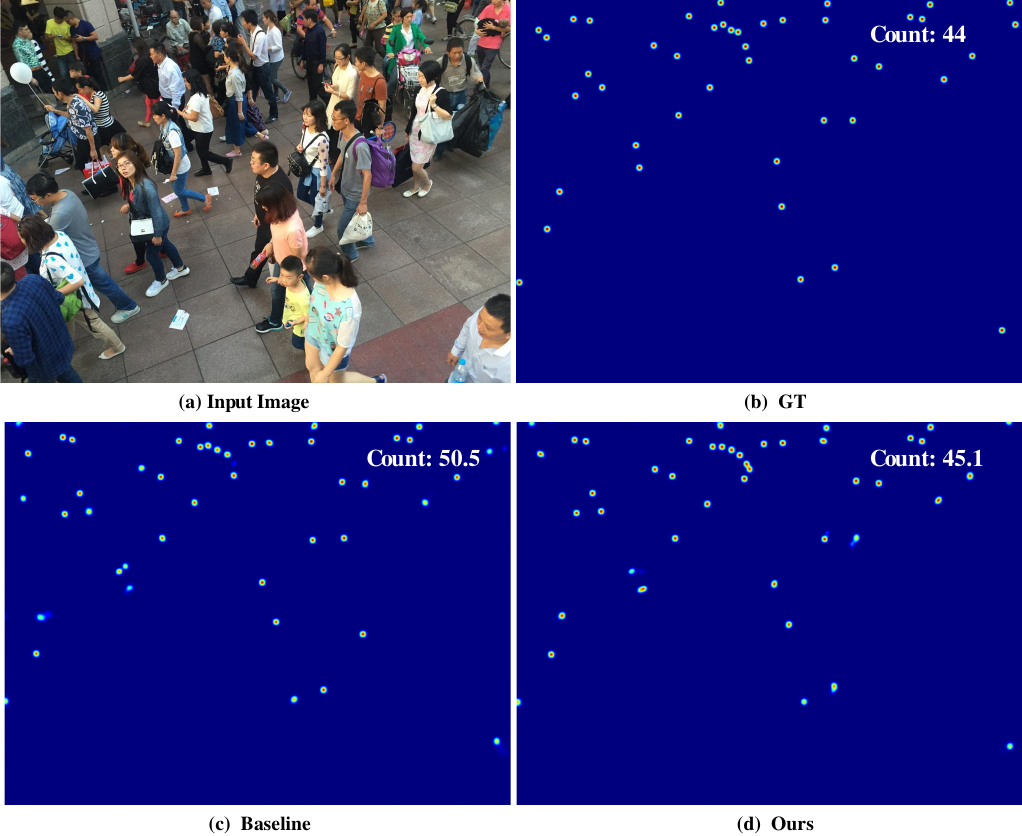}
    \caption{Visualization of predicted density maps from SHB.}
    \label{fig:density_map_shb}
\end{figure*}

\begin{figure*}[ht]
    \centering
    \includegraphics[width=1\textwidth]{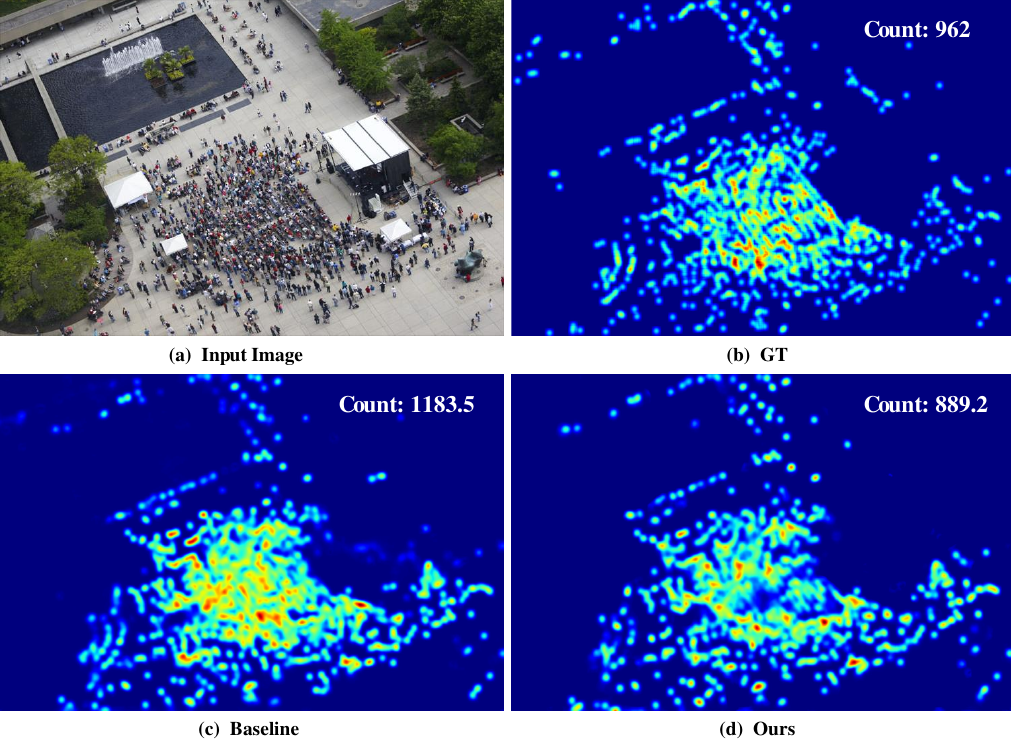}
    \caption{Visualization of predicted density maps from QNRF.}
    \label{fig:density_map_qnrf}
\end{figure*}

\begin{figure*}[ht]
    \centering
    \includegraphics[width=1\textwidth]{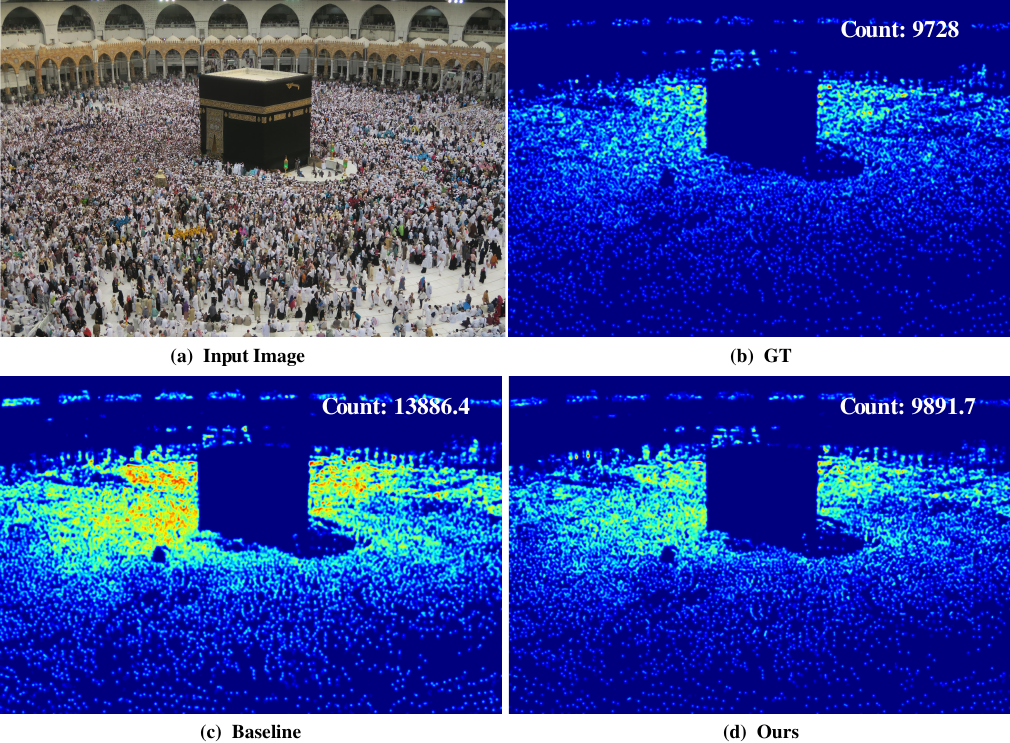}
    \caption{Visualization of predicted density maps from NWPU.}
    \label{fig:density_map_nwpu}
\end{figure*}

In the main manuscript, we have previously illustrated a selection of examples from the ShanghaiTech Part A (SHA) dataset. 
We now expand on this by presenting additional visual results derived from ShanghaiTech Part A (SHA), ShanghaiTech Part B (SHB), UCF-QNRF (QNRF), and NWPU Crowd (NWPU) datasets, corresponding to their respective test samples.
As can be observed in Figures~\ref{fig:density_map_sha}~\ref{fig:density_map_shb}~\ref{fig:density_map_qnrf}~\ref{fig:density_map_nwpu}, mPrompt$_{\ddagger}$ consistently outperforms mPrompt$_{reg}$ in generating superior density maps.
This superiority is apparent across various regions, whether dense or sparse, in each of the SHA, SHB, QNRF, and NWPU datasets. Thus, mPrompt$_{\ddagger}$ demonstrates marked improvement in performance across different types of crowd scenes.


\end{document}